\documentclass[a4paper, 12 pt, twocolumn, final, conference]{ieeetran}
\IEEEoverridecommandlockouts

\usepackage{graphicx}
\usepackage{subcaption}
\usepackage{times}   
\usepackage{amsmath} 
\usepackage{amssymb} 
\usepackage{avldefs} 
\usepackage{algorithm}
\usepackage{amsfonts}
\usepackage{algorithm, algorithmic}
\usepackage{array}
\usepackage{multirow}
\usepackage{microtype}
\usepackage{graphics}
\usepackage{booktabs} 
\usepackage{url}
\usepackage{bm}  

\usepackage[colorinlistoftodos,prependcaption,textsize=tiny]{todonotes}

\usepackage[square, numbers, compress]{natbib}

\usepackage{color}

\usepackage[UKenglish]{isodate}

\usepackage{booktabs}
\usepackage[normalem]{ulem}
\useunder{\uline}{\ul}{}

\title{\LARGE \bf
Trading with the Momentum Transformer: \\An Intelligent and Interpretable Architecture
}

\author{
    \IEEEauthorblockN{Kieran Wood\IEEEauthorrefmark{1}, Sven Giegerich\IEEEauthorrefmark{2}, Stephen Roberts\IEEEauthorrefmark{1}, Stefan Zohren\IEEEauthorrefmark{1}}
    \IEEEauthorblockA{\IEEEauthorrefmark{1}Oxford-Man Institute
of Quantitative Finance, University of Oxford}
    \IEEEauthorblockA{\IEEEauthorrefmark{2}Oxford Internet Institute, University of Oxford}
    \thanks{\IEEEauthorrefmark{1}Kieran Wood is the corresponding author and can be contacted via email: kieran.wood@eng.ox.ac.uk.}
}

\makeatletter
\renewcommand*{\thetable}{\arabic{table}}
\renewcommand*{\thefigure}{\arabic{figure}}
\let\c@table\c@figure
\makeatother 

\usepackage{caption}
\begin{document}
\maketitle
\thispagestyle{plain}
\pagestyle{plain}
%

\begin{abstract}
We introduce the \textit{Momentum Transformer}, an attention-based deep-learning architecture, which outperforms benchmark time-series momentum and mean-reversion trading strategies. Unlike state-of-the-art Long Short-Term Memory (LSTM) architectures, which are sequential in nature and tailored to local processing, an attention mechanism provides our architecture with a direct connection to all previous time-steps. Our architecture, an attention-LSTM hybrid, enables us to learn longer-term dependencies, improves performance when considering returns net of transaction costs and naturally adapts to new market regimes, such as during the SARS-CoV-2 crisis. Via the introduction of multiple attention heads, we can capture concurrent regimes, or temporal dynamics, which are occurring at different timescales. The \textit{Momentum Transformer} is inherently interpretable, providing us with greater insights into our deep-learning momentum trading strategy, including the importance of different factors over time and the past time-steps which are of the greatest significance to the model.
\end{abstract}

\section{Introduction}
Time-series momentum (TSMOM) strategies \cite{TimeSeriesMomentum}, also known as trend following or `follow the winner' strategies, are based on the simple heuristic of going long (short) on assets with positive (negative) returns over some lookback window. It is an observed anomaly in asset prices, meaning it performs contrary to the notion of the Capital Asset Pricing Model (CAPM) \cite{CAPM}.
It has been observed that stocks with larger relative returns  over the past year tend to have higher average returns over the subsequent year \cite{bornholt2012failure},
contradicting the efficient market hypothesis. The momentum effect has been widely studied \cite{TimeSeriesMomentum, CenturyOfTrendFollowing, TwoCenturiesOfTrendFollowing, AHLMomentum}
and TSMOM strategies are a consistent component of managed futures or Commodity Trading Advisors (CTAs).
The standard approach involves quantifying the magnitude of trends
\cite{AHLMomentum}
and sizing traded positions accordingly.
The univariate TSMOM strategies we focus on differ from the cross-sectional approach 
\cite{CrossSectionalMomentum}
which studies the comparative performance of assets. This is achieved by ranking a portfolio of assets based on their relative historical performance, for example, buying the top decile of assets and selling the bottom decile at a given time step. 

\textit{Deep Momentum Networks} (DMNs) \cite{DeepMomentum, SlowMomFastRev} are a deep-learning framework which can learn to size both the trend and position in a data driven manner by directly optimising on the  Sharpe  ratio  of  the  signal, typically using a Long Short-Term Memory (LSTM)
\cite{TimeSeriesDeepLearning}
architecture, significantly outperforming classical approaches from approximately 2003 on-wards, when electronic trading was becoming more prominent \cite{SlowMomFastRev}. The LSTM, initially proposed to address the vanishing and exploding gradient problem,
is a special kind of Recurrent Neural Network (RNN)
\cite{TimeSeriesDeepLearning},
which is a class of artificial neural networks models where information can flow from one step to another. In addition to exploiting long term trends, DMNs have also been observed to simultaneously exploit localised price fluctuations with a fast mean-reversion strategy \cite{SlowMomFastRev}. Mean-reversion \cite{MeanReversion}
strategies assume losers (winners) over some time horizon window will be winners (losers) in the subsequent period and are known as `follow the loser' strategies.

We note that even DMNs have been under-performing in recent years, such as during the SARS-CoV-2 crisis, which can be attributed to the presence of nonstationarity or momentum turning points, where a trend breaks down \cite{SlowMomFastRev}. Whilst the LSTM is good at learning local patterns \cite{TimeSeriesDeepLearning}, this architecture is known to struggle with long term patterns and responding to significant events, such as a market crash, which we term as regime change. Due to its sequential nature and resetting mechanism, the LSTM has a tendency to `forget' information from prior to any regime change, limiting its ability to capture global temporal dynamics.

One proven approach to making an LSTM DMN model more robust to regime change, in a data-driven manner, is via the introduction of an online changepoint detection (CPD) \cite{SlowMomFastRev} module to our DMN pipeline. The CPD module uses a principled Bayesian Gaussian Process 
region switching approach \cite{GPChangepointsAndFaults}, which is robust to noisy inputs, to detect regime change. This approach helps the model to quickly and correctly identify regime change, then respond accordingly. This technique, however, only utilises localised information and is unable to draw upon useful information from past, potentially similar, regimes. Furthermore, it can still place too much emphasis on the fast-reverting regime, resulting in poor performance when considering returns net of transaction cost.  

In this paper, we introduce the \textit{Momentum Transformer}, a subclass of DMNs which incorporates attention mechanisms \cite{NeuralMachineTrnslation}. An attention mechanism is a key-value lookup based on a given query. Initially proposed as a sequence-to-sequence \cite{Seq2Seq} model, attention based Transformer \cite{AttentionIsAllYouNeed} architectures have led to state-of-the-art performance in diverse fields, such as of natural language processing, computer vision, and speech processing \cite{TransformersSurvey}. 

In time-series applications, an attention mechanism uses a learnable weight function to measure the importance of previous timestamps. Transformer architectures have 
recently been harnessed for time-series modelling \cite{EnhancingLocalityTS, TFT, Informer}; however, former work has predominantly focused on forecasting tasks which typically feature periodic components and a higher signal-to-noise ratio than financial time series.
Attention mechanisms are known to lead to improvements in learning long term dependencies, with the ability to attend to significant events and learn regime specific temporal dynamics \cite{TimeSeriesDeepLearning}. It is noted in \cite{SlowMomFastRev} there are different regimes which are occurring concurrently at different timescales, which we can capture via the introduction of multiple attention heads. Whilst we observe the introduction of an attention mechanism to help respond to regime change, similarly to the inclusion of a CPD module, we observe that our architecture and a CPD module perform well together, leading to superior returns.


An important consideration when trading with DMNs is to understand why the model selects a given position. While \cite{SlowMomFastRev} provides some insight into the concurrent slow momentum and fast reversion strategies, with regard to examining how the model adjusts its position sizing during different regimes, the original DMN model is largely a black box. 
One of the innovations of the TFT \cite{TFT} is that it is constructed using components which are inherently interpretable, with its Variable Selection Network (VSN) naturally providing a measure of variable importance, taking time ordering into account. It provides insight into how the model combines different classical TSMOM and Moving Average Convergence Divergence (MACD) indicator \cite{AHLMomentum} strategies at different times. Furthermore, the attention patterns, which focus on previous time-steps, show significant structure and splits the time series into regimes. At a given time-step, the model focuses on alike regimes and places significant importance on momentum turning points. 

\section{Attention}
The self-attention mechanism, which relates positions of a sequence to construct a representation, is at the core of all Transformer-based architectures assessed in this paper. In essence, the mechanism incorporates a learnable similarity score, $\alpha: \mathcal{X} \times \mathcal{X} \to [0,1]$, to assign a measure of importance to previous time-steps. It was noted by \cite{TransformerDissection} that attention can be viewed as linear smoothing with a kernel smoother \cite{AllNonparametricStatistics} over the inputs, with the kernel scores measuring similarity. A kernel smoother aims to capture important patterns in noisy data.

We define the probability function $p(\mathbf{x}_\kappa|\mathbf{x}_q)\in[0,1]$ of feature vector $\mathbf{x}_\kappa \in \mathcal{X}$ when querying feature vector $\mathbf{x}_q \in \mathcal{X}$, which can also be interpreted as the attention weight $\alpha(\cdot, \cdot)$, or similarity score. The probability function is,
\begin{equation}
\label{eqn:att_prob}
p(\mathbf{x}_\kappa | \mathbf{x}_q) = \frac{k(\mathbf{x}_q, \mathbf{x}_\kappa) }{\sum_{\mathbf{x}_\kappa^\prime \in M(\mathbf{x}_\kappa, S_{\mathbf{X}_\kappa}) } k(\mathbf{x}_q, \mathbf{x}_\kappa^\prime)},
\end{equation}
for kernel function $k : \mathcal{X} \times \mathcal{X} \to \mathbb{R}^+$ and set filtering function $M:\mathcal{X} \times \mathcal{S} \to \mathcal{S}$ where ${S_{\mathbf{X}_\kappa}} \in \mathcal{S}$ is the set of keys $\{\mathbf{x}_{\kappa_1}, \ldots, \mathbf{x}_{\kappa_T}\}$. In the context of time-series, we can prevent access to future time-steps, which is a process known as masking. We define attention as,
\begin{equation}
    \mathrm{Att}(\mathbf{x}_q; M(\mathbf{x}_q, S_{\mathbf{X}_\kappa})) = \mathop{\mathbb{E}}_{\mathbf{x}_\kappa \sim p(\mathbf{x}_\kappa|\mathbf{x}_q)}[v(\mathbf{x}_\kappa)],
\end{equation}
for value function $v: \mathcal{X} \to \mathcal{Y}$.

Self-attention relates positions of a single sequence, i.e. $\mathbf{x}_q\in S_{\mathbf{X}_\kappa}$. The Transformer in \cite{AttentionIsAllYouNeed} uses an asymmetric exponential kernel $k(\mathbf{x}_q, \mathbf{x}_\kappa) = \exp\left(\frac{1}{\sqrt{d_\text{att}}}\langle \mathbf{W}_q \mathbf{x}_q , \mathbf{W}_\kappa \mathbf{x}_\kappa \rangle\right)$ with $v(\mathbf{x}_\kappa)=\mathbf{W}_v \mathbf{x}_\kappa$, where $d_\text{att}$ is the dimension we project $\mathbf{x}_q$ and $\mathbf{x}_\kappa$ into and dot-product $\langle\cdot,\cdot\rangle$ measures similarity. All weight matrices $\mathbf{W}_q\in\mathbb{R}^{d_\text{att} \times d_q}$, $\mathbf{W}_\kappa\in\mathbb{R}^{d_\text{att} \times d_q}$ and $\mathbf{W}_v\in\mathbb{R}^{d_\text{att} \times d_q}$ are learnable. We use the asymmetric exponential kernel in this paper for consistency with the literature \cite{AttentionIsAllYouNeed, EnhancingLocalityTS, Informer, TFT}, however, \cite{TransformerDissection} examines the usage of different kernels, including symmetric kernels.

\begin{figure}[tbph]
\centering

\includegraphics[width=1\linewidth]{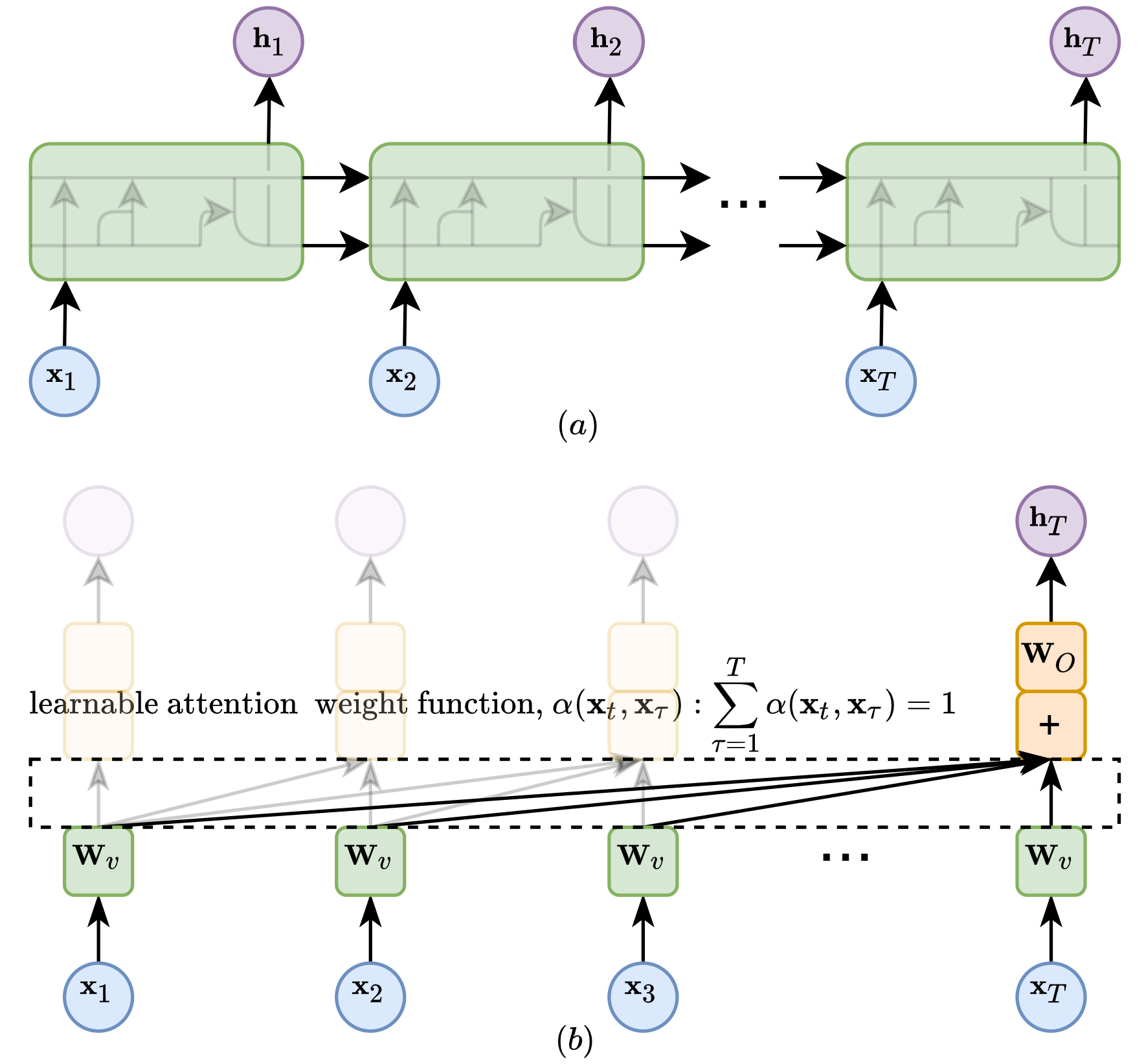}
\caption{(a) is an LSTM unrolled, demonstrating the sequential nature of the architecture (b) is an attention mechanism, demonstrating the direct link with each of the past time-steps. 
\label{fig:lstm-att-comparison}}
\end{figure}

The LSTM \cite{TimeSeriesDeepLearning}, detailed in Appendix \ref{apdx:additional_details}, is better suited than conventional RNNs for time-series forecasting problems with long sequences because, in addition to each hidden state $\mathbf{h}_t$ output, it maintains a cell state $\mathbf{c}_t$ for each time-step, which stores long-term information, modulated through a series of gates. The LSTM incorporates forget and input gates to help handle non-stationarity via a dynamic autocovariance structure.  However, due to the sequential nature of the architecture, demonstrated in Exhibit \ref{fig:lstm-att-comparison},  the LSTM is still inherently prone to forgetting when the sequence is large. The recursive structure can potentially lead to large error accumulations over long forecasting horizons and its resetting mechanism prevents access to information from similar regimes in the past. The attention mechanism forms a direct connection with each timestamp, alleviating both of these issues, meaning it is more appropriate for capturing long-term dependencies. 

As proposed in \cite{AttentionIsAllYouNeed}, we can expand the idea of attention to multiple heads, for different representation subspaces, where we learn optimal features at multiple scales and translations, resulting in improved learning capacity. Typically a larger kernel `bandwidth' results in greater averaging and hence smoother attention patterns. For each of the $H$ heads, we learn an attention weight function $\alpha_i: \mathbb{R}^{d_q} \times \mathbb{R}^{d_q} \to [0,1]$, where $i\in\{1,\ldots,H\}$, each corresponding to an instance of probability function \eqref{eqn:att_prob}, and value matrix $\mathbf{W}_{v, i} \in\mathbb{R}^{d_\text{att}\times d_q}$.  For time series $\mathbf{X}=\left(\mathbf{x}_t\right)_{t=1}^T$, multi-head attention (MHA) \cite{AttentionIsAllYouNeed} is defined as,
\begin{equation}
\label{eqn:mha}
\mathrm{MHA}(\mathbf{x}_t) = \sum_{i=1}^H(\mathbf{W}_{O,i})^\top\sum^T_{\tau=1} s_{t,\tau} \alpha_i(\mathbf{x}_t, \mathbf{x}_{\tau}) \mathbf{W}_{v, i} \mathbf{x}_{\tau}, 
\end{equation}
where we typically set $d_\text{att}=d_q/h$ and $\mathbf{W}_{O,i}, \in\mathbb{R}^{d_\text{att} \times d_q}$ is an additional linear mapping which projects each head back into $\mathbb{R}^{d_q}$. To ensure we only attend to previous time-steps, we can use masked multi-head attention (MMHA) with $s_{t,\tau} \in \{0,1\}$, which we can set to 0 for $\tau>t$. We now introduce cross-attention, a slight variant of \eqref{eqn:mha}. For query $\tilde{\mathbf{z}}_t$, given access to some representation $\mathbf{Y}=\left(\mathbf{y}_t\right)_{t=1}^T$, cross-attention (XA) is defined as
\begin{equation}
\label{eqn:cross_att}
\mathrm{XA}_\mathbf{Y}(\tilde{\mathbf{z}}_t) = \sum_{i=1}^H(\mathbf{W}_{O,i})^\top\sum_{\mathbf{y}_\tau \in S_\mathbf{Y}}  \alpha_i(\tilde{\mathbf{z}}_t, \mathbf{y}_{\tau}) \mathbf{W}_{v, i} \mathbf{y}_{\tau}, 
\end{equation}
 which uses the representation set $S_\mathbf{Y}$ for the keys and values, hence attending to the important information from the representation.


\section{Transformer Architectures}
In this section we present a simplified version of each of the Transformer architectures we assess, focusing on the key components and providing further details in Appendix \ref{apdx:additional_details}. The canonical Transformer \cite{AttentionIsAllYouNeed}, an encoder-decoder architecture, is a sequence-to-sequence model which first maps input sequence $\mathbf{X}=\left(\mathbf{x}_t\right)_{t=1}^T$ to encode an intermediate sequence of abstract representations, $\mathbf{Y}=\left(\mathbf{y}_t\right)_{t=1}^T$ with,
\begin{equation}
\label{eqn:trans_enc}
\mathrm{Enc}(\mathbf{X})  = (\mathrm{FFN} \circ \mathrm{MHA})(\mathbf{X}),
\end{equation}
where $\mathrm{MHA}$ is applied element-wise and $\mathrm{FFN}$ denotes a time distributed feed-forward network (FFN), applied to each position separately and identically. The FFN consists of a learnable linear transformation, or dense layer, followed by a a non-linear activation function $\max(\cdot,\mathbf{0})$, then another learnable linear transformation \cite{DeepLearningBook}. We can stack $M$ encoders with $(\mathrm{Enc}_1 \circ \ldots \circ \mathrm{Enc}_M)(\mathbf{X})$ to learn more complex representations. The decoder operates similarly on the output features, but with MMHA, to avoid look-ahead bias, followed by a cross-attention step, using the encoder representation $S_\mathbf{Y}$ for the keys and values. This attends to the important encoder information. We summarise the decoder as,
\begin{equation}
\label{eqn:decoder}
\mathrm{Dec}_\mathbf{Y}(\tilde{\mathbf{Z}})  = (\mathrm{FFN} \circ \mathrm{XA}_\mathbf{Y} \circ \mathrm{MMHA})(\tilde{\mathbf{Z}}),
\end{equation}
which we can again stack $M$ times. If our target decoder sequence is only of length one, we exclude MMHA in \eqref{eqn:decoder}. It has been demonstrated by \cite{EnhancingLocalityTS} that the decoder side of the transformer architecture can be sufficient for time series forecasting, which we refer to as the Decoder-Only Transformer, where we remove the encoder and the cross-attention piece in \eqref{eqn:decoder}. 

Input features $\mathbf{u}_t \in \mathcal{U}$ must first be converted to an embedding vector, or latent representation, $\mathbf{x}_t\in \mathbb{R}^{d_q}$. While RNN models capture the positional time-series pattern via their recurrent structure, the Transformer needs to preserve the positional context explicitly because the dot-product operation cannot capture this local context. We must inject some information about the relative or absolute position of the sequence items into the embedding which we detail in Appendix \ref{apdx:additional_details}. Whilst \cite{AttentionIsAllYouNeed} proposes `attention is all you need', doing away with convolutions and RNNs, \cite{TFT} suggests that it can still be beneficial to deal with positional encoding via an LSTM encoder for time-series applications. The TFT is, an attention-LSTM hybrid model which uses recurrent LSTM layers for local processing and self-attention layers for long-term dependencies.

It was proposed in \cite{TFT} that one can share the value weights in \eqref{eqn:mha} across heads, meaning each head can learn different temporal patterns while attending to a common set of input features. This is beneficial when interpreting the attention patterns and is referred to as Masked Interpretable MHA (MIMHA) where we replace $\mathbf{W}_{v, i}$ with $\frac{1}{h}\mathbf{W}_{v}$ in \eqref{eqn:mha}, sharing $\mathbf{W}_{v}$ across all heads. Additionally, The TFT also uses a sample-dependent Variable Selection Network (VSN) to filter out any inputs with a low signal rate, keeping only features which are of most significance for the prediction problem. Where $j\in\{1, \ldots,m\}$, $\tilde{\mathbf{x}}_{t, j}\in \mathbb{R}^{d_q}$ denotes the time-dependent embeddings of the covariates $\mathbf{u}_t\in\mathcal{U}$ our VSN output is,
\begin{equation}
    \mathbf{x}_t = \sum^{m}_{j=1}\eta(\tilde{\mathbf{x}}_{t,j})\psi_j(\tilde{\mathbf{x}}_{t,j}), \
    \text{s.t.} \ \sum^{m}_{j=1} \eta(\tilde{\mathbf{x}}_{t,j}) = 1
\end{equation}
where $\eta:\mathbb{R}^{d_q}\to[0,1]$ is a learnable function, of which we can interpret as the weighting for the $j^\text{th}$ covariate, and $\psi_j: \mathbb{R}^{d_q} \to \mathbb{R}^{d_q}$ is a learnable non-linear network specific to each covariate. A simplified model of a Decoder-Only TFT is, 
\begin{equation}
\label{eqn:tft_architecture}
\mathrm{TFT}(\tilde{\mathbf{X}})  = (\mathrm{FFN} \circ \mathrm{MIMHA} \circ \mathrm{LSTM} \circ \mathrm{VSN})(\tilde{\mathbf{X}}).
\end{equation}
Full details of the implementation can be found in Appendix \ref{apdx:additional_details}. Crucially, the architecture is designed with learnable skip components and optional non-linear processing, which means the model can become more complex, but only when necessary.

It has been noted that self-attention is typically sparse \cite{EnhancingLocalityTS, Informer}. Both the Convolutional Transformer \cite{EnhancingLocalityTS}, which is a variant of the Decoder-Only Transformer, and the Informer \cite{Informer}, which is a variant of the Transformer, modify the attention mechanisms in attempt to address this sparsity. Furthermore, the Convolutional Transformer introduces surrounding context into the attention mechanism and the Informer reduces the parameter space with each layer, via a process termed `distilling'.  We provide further details of each architecture in Appendix \ref{apdx:additional_details}.

\section{Momentum Transformer}
For all Transformer architectures tested, we adhere to the \textit{Deep Momentum Network} (DMN) framework \cite{DeepMomentum}. Our portfolio construction is typical of CTAs and the TSMOM literature \cite{TimeSeriesMomentum, TSMomAndVolScaling}. We use return data, which linearly detrends the price time-series. Because volatility varies across assets and time, an important part of the TSMOM framework is volatility scaling 
\cite{TSMomAndVolScaling, VolTargeting}, 
where we scale the returns of each asset by its volatility, to ensure that each asset has a similar contribution to the overall portfolio returns. It is another tool for ensuring approximate stationarity within regimes and increases leverage. We target an annualised volatility $\sigma_{\mathrm{tgt}}$, which we choose to be $15\%$ for consistency with previous works \cite{TimeSeriesMomentum, DeepMomentum, SlowMomFastRev}. The realised return of our strategy from day $t$ to $t+1$ is,
\begin{equation}
\label{eqn:tsmom}
R_{t+1}^\mathrm{TSMOM} = \frac{1}{N} \sum_{i=1}^{N} R_{t+1}^{(i)}, \quad R_{t+1}^{(i)} = z_t^{(i)}~\frac{\sigma_{\mathrm{tgt}}}{\sigma_t^{(i)}}~r_{t+1}^{(i)},
\end{equation}
where $N$ is the number of assets. For the $i$-th asset, $z_t^{(i)}$ is our position size and $\sigma_t^{(i)}$ the ex-ante volatility, calculated using a 60-day exponentially weighted moving standard deviation, which is in line with the literature \cite{TimeSeriesMomentum, TSMomAndVolScaling, DeepMomentum}. This approach is suited to a portfolio with a covariance matrix which is approximately diagonal. In this paper, we focus on trading futures, where there is substantially less covariance structure than equities, therefore, this construction is sufficient and removes complexity. Furthermore, we chose to remain consistent with the TSMOM literature, thus we can showcase the performance gains of our deep-learning approach.

 In the work by \cite{MomentumCrashes}, it is noted that momentum strategies work well until they don't, where they perform extremely poorly, and this is the key focus of our paper. Whilst there are lengthy periods of (approximate) stationarity in time-series constructed from historical futures data, the work by \cite{SlowMomFastRev} demonstrates, via a changepoint disequillibrium score, there are periods of significant non-stationarity, which can, relate to an, often abrupt, change in volatility, correlation length, mean-reversion length, or a combination.

The DMN framework uses deep-learning to simultaneously learn trend and size a position accordingly as,
\begin{equation}
    \mathbf{Z}_{T-\tau+1:T}^{(i)} = \left(\tanh  \circ f \circ g \right) \left(\mathbf{U}_{T-\tau+1:T}^{(i)}\right),
    \label{eqn:position-sizing}
\end{equation}
where $\mathbf{U}_{T-\tau+1:T}^{(i)}=(\mathbf{u}^{(i)}_t)^{T}_{t=T-\tau+1}$ is our series of input features, $\tau$ is our sequence length, $g(\cdot)$ is our candidate machine learning architecture and $f(\cdot)$ a time distributed, fully-connected layer, followed by an element-wise $\tanh$ activation function. For each time-step in the sequence, this maps next-day position as $z_t^{(i)} \in (-1,1)$, where $z^{(i)}_t = 1$ indicates a maximum long position and $z^{(i)}_t = -1$ a maximum short position. We train via mini-batch Stochastic Gradient Descent (SGD) \cite{DeepLearningBook},
with loss function selected to directly maximise some risk-adjusted metric, which for maximising Sharpe ratio involves minimising,
\begin{equation}
    \label{eqn:sharpe-loss}
    \mathcal{L}_{\mathrm{sharpe}} (\Omega; \, \pmb{\theta})  =  -  \frac
    { \sqrt{252} \, \mathbb{E}_\Omega\left[R_t^{(i)}\right] }
    {\sqrt{ \mathrm{Var}_\Omega\left[R_t^{(i)}\right] }},
\end{equation}
where $\Omega$ contains all asset-time pairs in the mini-batch and $\pmb{\theta}$ is the vector of all trainable parameters.

\begin{figure}[tbph]
\centering
\includegraphics[width=0.75\linewidth]{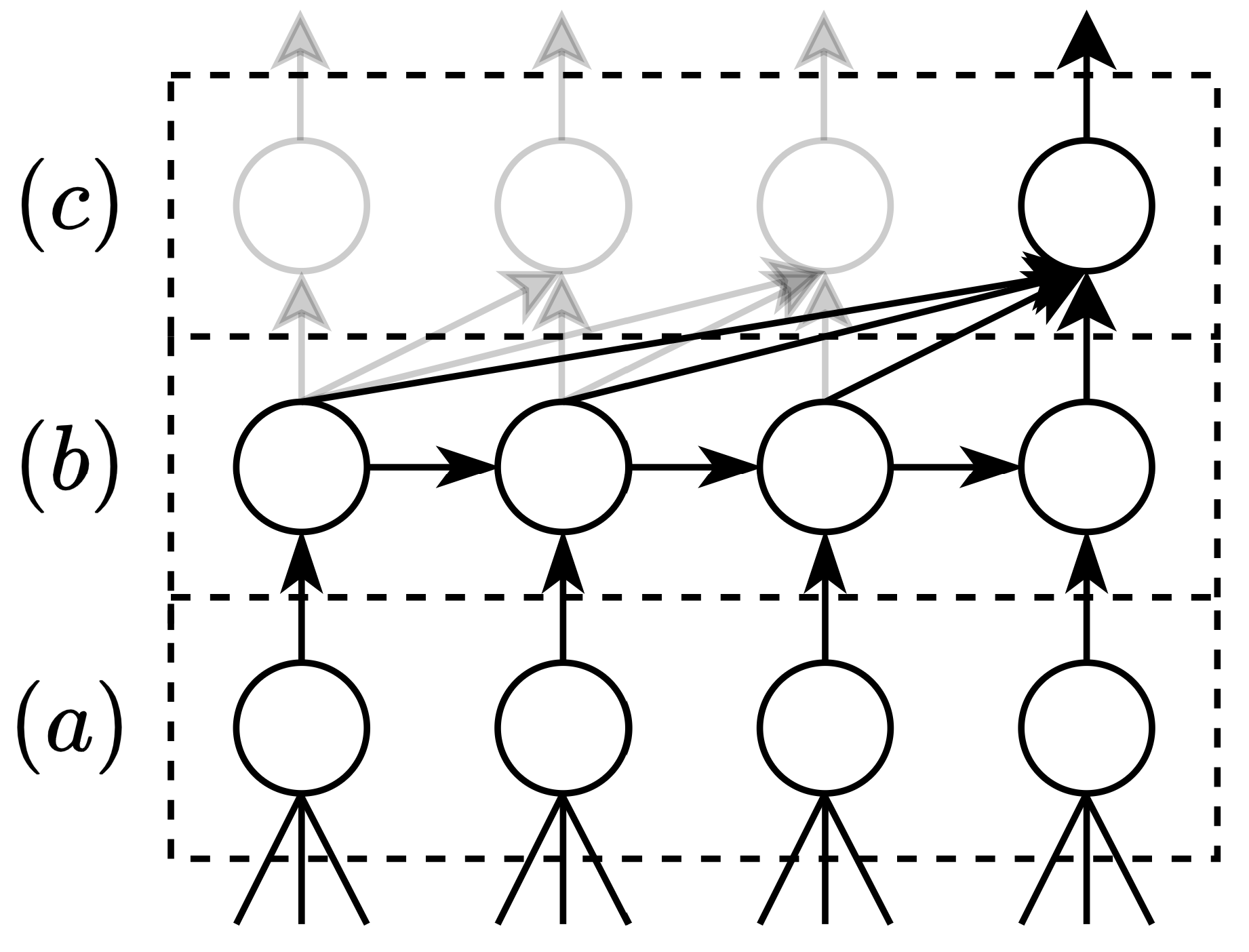}
\label{fig:simple-mom-trans}
\caption{A (simplified) \textit{Momentum Transformer} architecture, corresponding to $g(\cdot)$, pieces together (a) Variable Selection Network, (b) LSTM, and (c) self-attention mechanism.
}
\end{figure}

Our input features, which are common signals used in the TSMOM literature \cite{TimeSeriesMomentum, AHLMomentum}, include returns at different timescales $\hat{r}_{(\cdot)}^{(i)}$, corresponding to daily, monthly, quarterly, biannual and annual returns, which are normalised using ex-ante volatility $\sigma_t^{(i)}$. We also use MACD \cite{AHLMomentum} indicators which are a volatility normalised moving average convergence divergence indicator $\mathrm{M}_{t}^{(i)}(S,L)$, defining the relationship between a short $S$ and long signal $L$. The implementation of MACD indicators in DMNs is detailed in \cite{DeepMomentum}. 

We can optionally incorporate a CPD module, which is a feature preprocessing step for each time $t$, where a single changepoint is assumed in a look-back window of length $l$. The additional features we add for each time-step are changepoint severity $\nu^{(i)}_t(l)\in(0,1)$, where a high severity score suggests regime change over lookback window $l$ is highly likely, and changepoint location $\gamma^{(i)}_t(l)\in(0,1)$, which measures the normalised location of the changepoint in the window. Details can be found in \cite{SlowMomFastRev}, however, we perform CPD across different timescales $l\in \{21, 126\}$ , of a month and half-year. Multiple CPD timescales are enabled by the VSN component of the TFT, which removes any unnecessary inputs which negatively impact performance.  At longer timescales the CPD module focuses on larger, more significant events and at shorter timescales, our model can detect events more rapidly or exploit more localised fluctuations.


Further details of the implementation are provided in Appendix \ref{apdx:additional_details}. Sample code for the \textit{Momentum Transformer} is available\footnote{\url{https://github.com/kieranjwood/trading-momentum-transformer}}.

\section{Back-testing Details}
For all of our experiments, we used a portfolio of 50 of the most liquid, continuous futures contracts over the period 1990--2020, extracted from the Pinnacle Data Corp CLC Database\footnote{\url{https://pinnacledata2.com/clc.html}}. It is a balanced portfolio consisting  of commodities (CM), equities (EQ), fixed income (FI) and foreign exchange (FX) futures. The futures dataset is backwards ratio adjusted to create a continuous price series. Further details can be found in Appendix \ref{apdx:data}. This dataset has been previously used to benchmark strategies
\cite{SlowMomFastRev, DeepReinforcementLearning}.

We use an expanding window approach, where we initially use 1990--1995 for training-validation, then test out-of-sample on the period 1995--2000, expand the training-validation set to 1990--2000, test out-of-sample on the subsequent five years, and so on. We present results for three different scenarios:
\begin{enumerate}
    \item \textbf{Average test results over all five year windows 1995--2020}, allowing us to measure performance over a sustained period and additionally capture how effective the architecture is early on, when limited data is available. 
    \item \textbf{Test results over the period 2015--2020}, which provides insight into how our candidates perform across recent years which exhibited significant nonstationarity. In this time period both classical strategies and LSTM-based DMNs have been observed to under-perform, however CPD has previously been observed to somewhat alleviate this degradation \cite{SlowMomFastRev}.
    \item \textbf{The SARS-CoV-2 crisis}, from 1 January 2020 until 15 October 2020, to observe how our candidate architectures deals with regime change, including the market crash and the subsequent Bull market.
\end{enumerate}

Our \textit{Momentum Transformer} candidate architectures correspond to $g(\cdot)$ in Equation \eqref{eqn:position-sizing} and were chosen as: 1) Transformer, 2) Decoder-Only Transformer, 3) Convolutional Transformer, 4) Informer, and 5) Decoder-Only TFT. Additionally, we tested a variation of the best performing architecture, the Decoder-Only TFT, where we included CPD covariates. All details of the experiments can be found in Appendix \ref{apdx:experiment_settings}.

\begin{table*}[htbp]
\centering
\caption{Strategy Performance Benchmark -- Raw Signal Output.}
\label{tbl:results-raw}
\begin{tabular}{lrrrrrrrrr}
\toprule
\textbf{}                 & \textbf{Returns} & \textbf{Vol.} & \textbf{Sharpe}   & \textbf{\begin{tabular}[c]{@{}l@{}}Down.\\     Dev.\end{tabular}}  & \textbf{Sortino} & \textbf{MDD}    & \textbf{Calmar} & \textbf{\begin{tabular}[c]{@{}l@{}}\% $+$ve \\     Returns\end{tabular}} & \textbf{$\mathbf{\frac{\text{Ave.  P}}{\text{Ave. L}}}$} \\ \midrule
{\ul \textbf{Average 1995--2020}} \\
Long-Only            &         2.45\% &             4.95\% &         0.51 &         3.51\% &          0.73 &       12.51\% &         0.21 &          52.43\% &             0.988 \\
TSMOM                &         4.43\% &             4.47\% &         1.03 &         3.11\% &          1.51 &        6.34\% &         0.94 &          54.23\% &             1.002 \\
LSTM                 &         2.71\% &             1.67\% &         1.70 &         1.10\% &          2.66 &        2.14\% &         1.68 &          55.17\% &             1.091 \\
Transformer           &         3.14\% &             2.49\% &         1.41 &         1.68\% &          2.13 &        2.92\% &         1.53 &          54.71\% &             1.051 \\
Decoder-Only Trans.  &         2.95\% &             2.61\% &         1.11 &         1.74\% &          1.69 &        3.47\% &         1.09 &          53.50\% &             1.051 \\
Conv. Transformer    &         2.94\% &             2.75\% &         1.07 &         1.87\% &          1.60 &        3.80\% &         0.98 &          53.55\% &             1.041 \\
Informer             &         2.39\% &             1.38\% &         1.72 &         0.89\% &          2.67 &        1.43\% &         1.79 &          54.88\% &             1.103 \\
Decoder-Only TFT     &         \textbf{4.01}\% &             1.54\% &         2.54 &         0.96\% &          4.14 &        1.32\% &         \textbf{3.22} &          57.34\% &             \textbf{1.154} \\
Decoder-Only TFT CPD &         3.70\% &             \textbf{1.37}\% &         \textbf{2.62} &         \textbf{0.85\%} &          \textbf{4.25} &        \textbf{1.29\%} &         \textbf{3.22} &          \textbf{57.66}\% &             1.151 \\ \midrule
{\underline{\textbf{Average 2015--2020}}} \\
Long-Only            &         1.73\% &             5.00\% &         0.37 &         3.59\% &          0.51 &       11.41\% &         0.15 &          51.97\% &             0.982 \\
TSMOM                &         0.97\% &             4.38\% &         0.24 &         3.19\% &          0.33 &        8.25\% &         0.12 &          52.82\% &             0.931 \\
LSTM                 &         1.23\% &             1.85\% &         0.82 &         1.32\% &          1.19 &        3.55\% &         0.66 &          53.38\% &             1.004 \\
Transformer          &         1.98\% &             1.29\% &         1.53 &         0.85\% &          2.32 &        1.07\% &         1.86 &          54.76\% &             1.071 \\
Decoder-Only Trans.  &         1.37\% &             1.97\% &         0.72 &         1.37\% &          1.03 &        2.63\% &         0.60 &          52.76\% &             1.012 \\
Conv. Transformer    &         1.85\% &             1.92\% &         0.98 &         1.30\% &          1.47 &        3.14\% &         0.77 &          52.93\% &             1.056 \\
Informer             &         1.67\% &             1.09\% &         1.51 &         0.72\% &          2.30 &        1.17\% &         1.44 &          54.39\% &             1.089 \\
Decoder-Only TFT     &         1.99\% &             1.23\% &         1.71 &         0.82\% &          2.61 &        1.17\% &         2.06 &          55.72\% &             1.073 \\
Decoder-Only TFT CPD &         \textbf{2.06\%} &             \textbf{1.02\%} &         \textbf{2.00} &         \textbf{0.66\%} &          \textbf{3.10} &        \textbf{0.82\%} &         \textbf{2.53} &          \textbf{55.74\%} &             \textbf{1.120} \\ \midrule
{\underline{\textbf{SARS-CoV-2}}} \\
Long-Only            &                  -1.46\% &                       6.73\% &                  -0.19 &                   5.64\% &                   -0.22 &                 12.32\% &                  -0.12 &                    57.28\% &                       0.720 \\
TSMOM                &                   0.90\% &                       4.73\% &                   0.21 &                   3.14\% &                    0.32 &                  4.17\% &                   0.22 &                    50.00\% &                       1.041 \\
LSTM                 &                  -4.15\% &                       2.82\% &                  -1.50 &                   2.52\% &                   -1.67 &                  5.35\% &                  -0.78 &                    52.29\% &                       0.643 \\
Transformer          &                   4.42\% &                       \textbf{1.28\%} &                   \textbf{3.38} &                   \textbf{0.83\%} &                    \textbf{5.55} &                  \textbf{0.84\%} &                   7.31 &                    \textbf{64.85\%} &                       1.066 \\
Decoder-Only Trans.  &                   \textbf{8.02\%} &                       2.58\% &                   3.01 &                   1.42\% &                    \textbf{5.55} &                  1.05\% &                   \textbf{8.56} &                    58.83\% &                       \textbf{1.243} \\
Conv. Transformer    &                   3.13\% &                       1.99\% &                   1.81 &                   1.40\% &                    2.74 &                  1.61\% &                   3.17 &                    57.48\% &                       1.058 \\
Informer             &                   4.30\% &                       1.60\% &                   2.71 &                   1.00\% &                    4.45 &                  1.07\% &                   4.28 &                    59.61\% &                       1.137 \\
Decoder-Only TFT     &                   1.81\% &                       1.75\% &                   1.22 &                   1.37\% &                    1.74 &                  2.14\% &                   1.57 &                    60.39\% &                       0.831 \\
Decoder-Only TFT CPD &                   3.39\% &                       1.51\% &                   2.47 &                   1.03\% &                    4.08 &                  1.15\% &                   5.92 &                    59.90\% &                       1.068 \\
\bottomrule
\end{tabular}

\end{table*}

\section{Results and Discussion}
\subsection{Performance}
We have recorded the results, for each of our three testing scenarios, in Exhibit \ref{tbl:results-raw}. We consider,
\begin{enumerate}
    \item \textbf{profitability} through annualised returns and percentage of positive captured returns,
    \item \textbf{risk} through annualised volatility, annualised downside deviation and maximum drawdown (MDD), and
    \item \textbf{risk-adjusted performance} through annualised Sharpe, Sortino and Calmar ratios.
\end{enumerate}
Since we completely rerun each experiment five times, we have reported the average across all runs, for each metric. The Decoder-Only TFT, which we will refer to as the \textit{Momentum Transformer}, outperforms the benchmark architectures across all risk-adjusted performance metrics for scenarios 1 and 2. Notably, compared to the LSTM, Sharpe ratio is improved by 50\% during the period 1995--2020 and during 2015--2020 the improvement is 109\%. In general, compared to the LSTM across all experiments, the \textit{Momentum Transformer} has higher returns, predicts the direction of price more often and reduces all risk metrics overall. Whilst a lookback of approximately one annual quarter has previously been found to be optimal for LSTM-based DMNs, we note that the \textit{Momentum Transformer} is able to learn longer-term patterns and works better with an input sequence length of one year.

\begin{figure*}[tbph]
\centering
\includegraphics[width=1\linewidth]{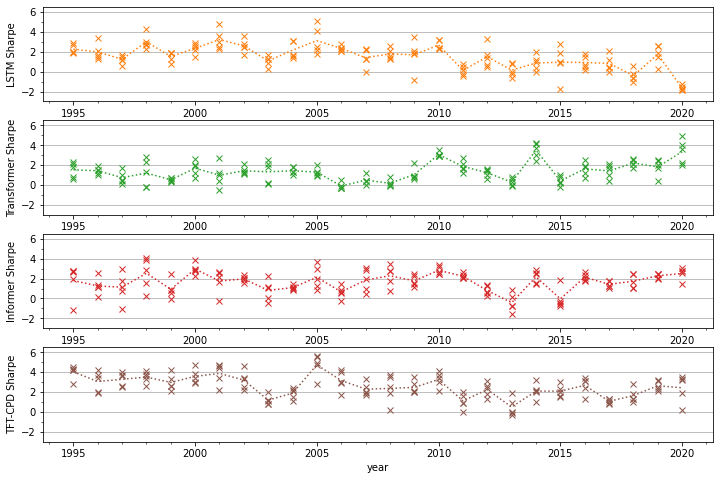}
\caption{Average annual Sharpe ratio by year, including the results for each of the five experiment repeats. \label{fig:gp-benchmark}} 
\end{figure*}
\begin{figure*}[tbph]
\centering
\includegraphics[width=1\linewidth]{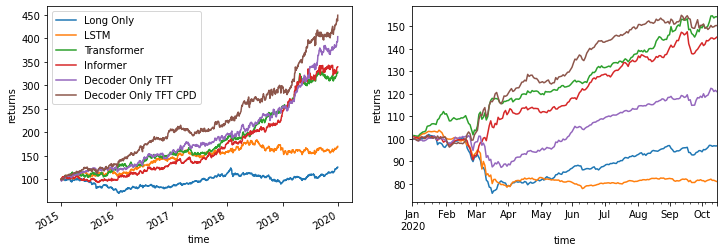}
\caption{These plots benchmark our strategy performance for the 2015--2020 scenario (left) and the SARS-CoV-2 scenario (right). For each plot we start with \$100 and we re-scale returns to 15\% volatility. Since we ran each experiment five times, we plot the repeat which resulted in the median Sharpe ratio, across the entire experiment. \label{fig:strat-benchmark}}
\end{figure*}

We demonstrate that the addition of a CPD module can be complementary, rather than an alternative to multi-head attention, and in the period 2015--2020 we observe a further improvement in Sharpe ratio of 17\% for the \textit{Momentum Transformer}. Furthermore, we demonstrated that it can be beneficial to input both a short LBW of one month and a longer LBW of half a year, allowing the Variable Selection Network to determine when these inputs are relevant in a data-driven manner. 

The work in \cite{DeflatedSharpe} explores the issue of back-test over-fitting, and how it can artificially inflate Sharpe ratio and propose that this may need to be corrected. With our expanding window approach, we are able to calculate out-of-sample Sharpe for all years from 1995 to 2021, shown in Exhibit \ref{fig:gp-benchmark}. Whilst the other transformer architectures do not perform as well as the Decoder-Only TFT in the first two experiments, Exhibit \ref{fig:gp-benchmark} reveals that there is actually a clear upward trend in recent years, during which the performance of the other non-hybrid Transformer architectures is comparable to the TFT. During the SARS-CoV-2 crisis the Transformer (canonical and Decoder-Only), and the Informer outperform even the Decoder-Only TFT. This period of clearly defined regimes, with a large crash followed by a Bull market, is unsurprisingly home turf for attention based architectures. The LSTM exhibits very poor performance during this experiment and we argue that the LSTM is better suited to exploiting short term patterns. In contrast, the LSTM still performs reasonably well during the 2008 financial crisis. This is likely because there was more signal in the lead up to this event, compared to the SAR-CoV-2 crash which was sudden and caused by exogenous factors. Being an attention-LTSM hybrid, the TFT model tends to be more of an all-rounder, with a more stable average Sharpe ratio across all years. It should be noted that we observe slightly more variance in the repeats of experiments for the TFT, which is likely attributed to the fact that the TFT is a more complex architecture and hence more sensitive to the tunable hyperparameters.

Interestingly, the canonical Transformer outperforms the Informer during the SARS-CoV-2 crisis and over the 2015--2020 period. While the Informer has proven to produce superior results for other applications \cite{Informer}, it is thus not necessarily the most suitable architecture for momentum trading. This could be because the Informer is designed for time-series which exhibit stronger periodicity or a setting with a higher signal-to-noise ratio in general. Similarly, the Convolutional Transformer does not outperform the Decoder-Only Transformer, again highlighting the challenges of attending to localised patterns in a low signal-to-noise setting.

We provide plots of returns for experiment scenarios 2 and 3 in Exhibit  \ref{fig:strat-benchmark}. These plots further illustrate how the LSTM is unsuitable during the market nonstationarity of 2015--2020 and during the SARS-CoV-2 crisis. Whilst the Transformer architectures are all able to respond naturally to sudden regime change, especially in comparison to the LSTM, we do observe that the addition of the CPD module still significantly helps with the timing of this response. It is evident that the non-hybrid Transformer models perform exceptionally well once the Bull market is established after the SARS-CoV-2 market crash, exploiting this regime with slow momentum.

\subsection{Interpretability}
Not only is the TFT-based architecture the best performing, but it also has additional benefits of being more interpretable. We analyse two components, both of which are detailed in \cite{TFT},
\begin{enumerate}
    \item \textbf{variable importance}, demonstrating how different classical strategies are blended at different times, in addition to their interaction with features from the CPD module, and
    \item \textbf{interpretable multi-head attention}, providing insight into how our model focuses on significant events and similar regimes.
\end{enumerate}
It should be noted that the LSTM forget gate can offer some insight into regime change by quantifying the magnitude of `forgetting' long term information; however, unlike the attention mechanism, it cannot provide insight into regimes other than the fact it is forgetting the one directly prior to the changepoint.

\begin{table}
\centering
\caption{Decoder-Only TFT average variable importance.}
\label{tbl:variable_importance}
\begin{tabular}{l|rr|rr}
\toprule
{} & \multicolumn{2}{c|}{\textbf{2015--2020}} & \multicolumn{2}{c}{\textbf{SARS-CoV-2}} \\
{} &    no CPD  & with CPD & no CPD  & with CPD \\
\midrule
$\hat{r}_\textrm{day}$     & 30.8\% & 23.3\% & 24.4\% & 21.2\% \\
$\hat{r}_\textrm{month}$   & 13.6\% &  5.7\% & 10.6\% &  7.4\% \\
$\hat{r}_\textrm{quarter}$ &  8.9\% &  7.4\% & 14.0\% &  5.7\% \\
$\hat{r}_\textrm{biannual}$   &  8.9\% &  6.2\% &  8.5\% &  7.5\% \\
$\hat{r}_\textrm{annual}$    & 11.9\% & 10.5\% & 13.5\% &  8.8\% \\
$\mathrm{M}_{t}^{(i)}(8,24)$             &  9.1\% &  7.3\% &  9.7\% & 12.9\% \\
$\mathrm{M}_{t}^{(i)}(16,48)$            & 10.3\% &  6.6\% & 11.9\% &  6.3\% \\
$\mathrm{M}_{t}^{(i)}(32,98)$            &  6.5\% &  5.7\% &  7.3\% &  8.7\% \\
CPD Score 21           &   - &  6.4\% &   - &  4.1\% \\
CPD LBW 21              &   - &  7.0\% &   - &  7.6\% \\
CPD Score 126           &   - &  4.7\% &   - &  4.8\% \\
CPD LBW 126              &   - &  9.2\% &   - &  4.9\% \\
\bottomrule
\end{tabular}
\end{table}

\begin{figure}[tbph]
\centering
\includegraphics[width=1\linewidth]{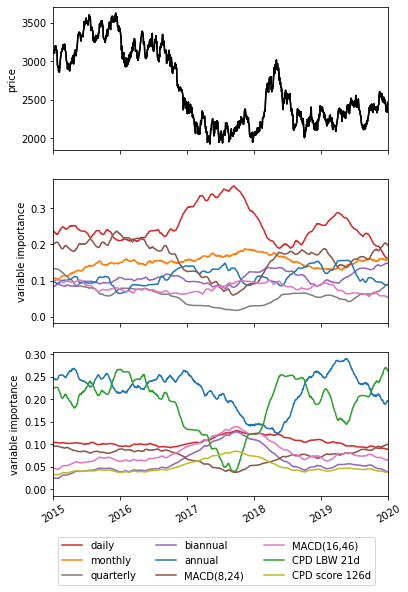}
\caption{Variable importance for Cocoa future, forecasting out-of-sample over the period 2015--2020. The middle plot is the Decoder-Only TFT and the bottom plot is the Decoder-Only TFT with CPD. For each model, we only plot the seven features with the highest average weighting. We plot Cocoa because it exhibits typical behaviour of the model trading a commodity future and is comprised of a series of clearly defined regimes during this period. \label{fig:variable-importance}}
\end{figure}

In Exhibit \ref{tbl:variable_importance} we tabulate the variable importance for the Decoder-Only TFT, averaged across 2015--2020 and then for the SARS-CoV-2 crisis. We record results for the models, with and without CPD, averaged over all repeats of the experiment. Overall, we note that the daily return feature is allocated the highest weighting, however this is somewhat reduced after the addition of CPD. This suggests that the model is relying less on fast reversion and, with a total weighting of 27\% allocated to the CPD features, it is apparent that the model is exploiting the CPD information. Interestingly CPD LBW length tends to be of greater importance than the score, indicating that the model is learning from patterns in this length metric. After the addition of CPD, the model relies less on return timescales other than the shortest (daily) and the longest (annual) timescale.

Comparatively, daily data information is less important during the SARS-CoV-2 crisis than the 2015--2020 period. This is likely because 2015--2020 is a highly non-stationary period, however 2020 is characterised by a large crash followed by a clear uptrend. During this period quarterly returns are given much more weighting, in the lack of CPD. Across the board, the MACD indicators are allocated above average importance, tweaking the importance of each depending on the scenario and whether it has access to CPD information. Interestingly, Exhibit \ref{tbl:variable_importance} shows that CPD features are not given a high importance during the SARS-CoV-2 experiment, despite their inclusion doubling the Sharpe ratio. This could be attributed to the fact that the feature is important for the crash, but then less important once a trend is established. In Exhibit \ref{fig:variable-importance} we can see how the variable importance for trading on Cocoa futures changes over time, where our model intelligently blends different strategies at different points in time, noting a change in strategy after significant events. We provide a detailed discussion of this in Appendix \ref{apdx:additional_discussion}.

\begin{figure}[tbph]
\includegraphics[width=\linewidth]{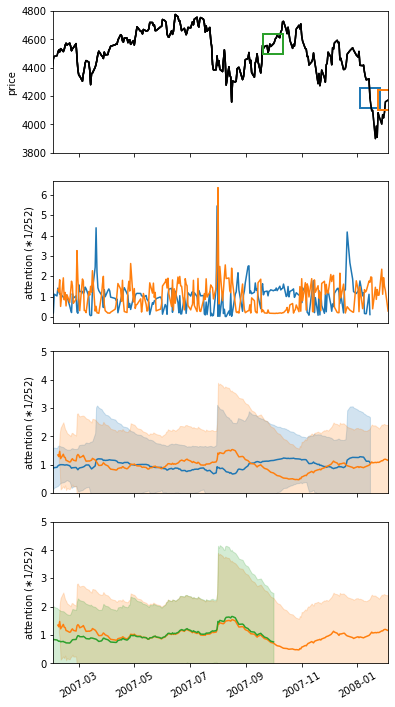}
\caption{We plot the attention pattern for a single run of our Decoder-Only TFT CPD model, forecasting out-of-sample on FTSE 100 data in the lead up to the 2008 financial crash. We plot the patterns for 1 October 2007 (green), 15 January 2008 (blue), and 2 April 2008 (orange). In each case we are running the model for the next day, therefore, the plots all correspond to the query at the most recent time-step. We take the average attention weight across heads. The third and fourth plots show the 21 day exponentially weighted rolling average and standard deviations, with the shaded area representing the 95\%-ile ($\pm$ 1.96 standard deviations).
These plots indicate how attention patterns are similar for points in the same type of regime, but oppose each other for different regimes. \label{fig:attention_ftse_2008}}
\end{figure}

\begin{figure}[tbph]
\centering

\includegraphics[width=1\linewidth]{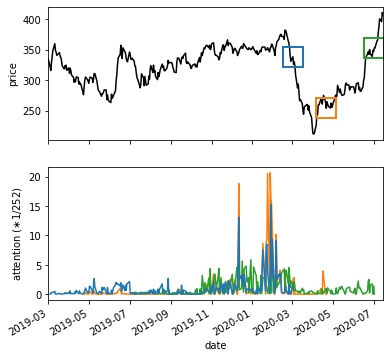}
\caption{Lumber future price during SARS-CoV-2 crisis and the associated attention pattern, with attention weight aggregated across heads, when making a prediction at 1 March 2020 (blue), 21 April 2020 (orange), and 2 July 2020 (green), indicating that significant attention is placed on momentum turning points. We plot this future because it is characterised by a sudden crash caused by exogenous factors, immediately followed by a skyrocketing price. 
\label{fig:attention_lumber_covid}
}
\end{figure}

The self-attention plots in Exhibits \ref{fig:attention_ftse_2008} and \ref{fig:attention_lumber_covid} both illustrate significant structure. Exhibit \ref{fig:attention_ftse_2008} demonstrates that greater attention is placed on similar regimes. Here two points in similar regimes, where there is a clear upward trend, have almost an identical pattern, only deviating from each other at the furthest time-steps. On the other hand, the attention pattern for a point inside a clear downward trend focuses more on other downward trends, and increases (decreases) its attention pattern when the other plots are decreasing (increasing). Approximately two-thirds along the plot, there appears to be some sort of stationary regime, where the opposing patterns stay approximately constant. Interestingly, the attention patterns tend to place significant attention on relevant momentum tuning points, partitioning the time-series into regimes and indicating that this is taken into account when selecting a strategy. 

Structure in the attention pattern is further demonstrated during the SARS-CoV-2 crisis, in Exhibit \ref{fig:attention_lumber_covid}, where our model recognises fundamental change caused by some exogenous factors. In this example, the peaks indicating momentum turning points are even more pronounced, clearly segmenting the plot into regimes. This highlights the significance our model places on momentum turning points when selecting a strategy. Again, different turning points, are of greater importance depending on the point of reference.

\subsection{Results Net of Transaction Costs}

\begin{table}[htbp]
\centering
\caption{Transaction cost impact on Sharpe Ratio over 2015--2020 for individual assets, averaged by asset class, and for the entire diversified portfolio.}
\label{tbl:cost-matrix-2015-2020}
\begin{tabular}{lrrrrrrr}
\toprule
$C$ bps &  0.0 &  0.5 &  1.0 &  1.5 &  2.0 &  2.5 &  3.0 \\
\midrule
\multicolumn{5}{l}{{\underline{\textbf{LSTM}}}} \\
CM         &              0.12 &                      0.09 &                      0.05 &                      0.01 &                     -0.02 &                     -0.06 &                     -0.10 \\
EQ         &              \textbf{0.37} &                      0.32 &                      0.27 &                      0.22 &                      0.16 &                      0.11 &                      0.06 \\
FI         &              0.09 &                     -0.11 &                     -0.32 &                     -0.53 &                     -0.74 &                     -0.94 &                     -1.15 \\
FX         &              0.11 &                      0.01 &                     -0.08 &                     -0.18 &                     -0.27 &                     -0.37 &                     -0.46 \\
Port.       &              0.82 &                      0.51 &                      0.20 &                     -0.12 &                     -0.43 &                     -0.74 &                     -1.05 \\
\midrule 
\multicolumn{5}{l}{{\underline{\textbf{Transformer}}}} \\
CM         &              0.27 &                      0.23 &                      0.19 &                      0.15 &                      0.11 &                      0.08 &                      0.04 \\
EQ         &              \textbf{0.37} &                      \textbf{0.33} &                      \textbf{0.28} &                      \textbf{0.23} &                      \textbf{0.19} &                      \textbf{0.14} &                      \textbf{0.10} \\
FI         &              0.23 &                      0.03 &                     \textbf{-0.16} &                     \textbf{-0.35} &                     \textbf{-0.55} &                     \textbf{-0.74} &                     \textbf{-0.93} \\
FX         &             -0.17 &                     -0.24 &                     -0.32 &                     -0.39 &                     -0.47 &                     -0.55 &                     -0.62 \\
Port.       &              1.53 &                      1.26 &                      0.99 &                      0.72 &                      \textbf{0.45} &                      \textbf{0.18} &                     \textbf{-0.09} \\
 \midrule
\multicolumn{5}{l}{{\underline{\textbf{Informer}}}} \\
CM         &              0.28 &                      0.24 &                      0.19 &                      0.15 &                      0.10 &                      0.06 &                      0.01 \\
EQ         &              0.34 &                      0.28 &                      0.22 &                      0.16 &                      0.10 &                      0.04 &                     -0.02 \\
FI         &              0.08 &                     -0.13 &                     -0.35 &                     -0.56 &                     -0.78 &                     -0.99 &                     -1.20 \\
FX         &             -0.14 &                     -0.24 &                     -0.33 &                     -0.43 &                     -0.53 &                     -0.62 &                     -0.72 \\
Port.       &              1.51 &                      1.17 &                      0.83 &                      0.49 &                      0.15 &                     -0.19 &                     -0.53 \\
 \midrule
\multicolumn{5}{l}{{\ul \textbf{Decoder-Only TFT}}}  \\
CM         &              0.44 &                      0.40 &                      0.35 &                      0.31 &                      0.26 &                      0.22 &                      0.17 \\
EQ         &              0.25 &                      0.19 &                      0.13 &                      0.07 &                      0.02 &                     -0.04 &                     -0.10 \\
FI         &              \textbf{0.30} &                      \textbf{0.05} &                     -0.20 &                     -0.45 &                     -0.69 &                     -0.94 &                     -1.18 \\
FX         &              \textbf{0.28} &                      \textbf{0.18} &                      \textbf{0.08} &                     \textbf{-0.02} &                     \textbf{-0.12} &                     \textbf{-0.22} &                     \textbf{-0.32} \\
Port.       &              1.71 &                      1.36 &                      1.01 &                      0.67 &                      0.32 &                     -0.03 &                     -0.37 \\ \midrule
\multicolumn{5}{l}{{\underline{\textbf{Decoder-Only TFT CPD}}}} \\
CM         &              \textbf{0.55} &                      \textbf{0.50} &                      \textbf{0.45} &                      \textbf{0.40} &                      \textbf{0.35} &                      \textbf{0.30} &                      \textbf{0.25} \\
EQ         &              0.18 &                      0.12 &                      0.05 &                     -0.01 &                     -0.07 &                     -0.14 &                     -0.20 \\
FI         &              0.23 &                     -0.03 &                     -0.29 &                     -0.55 &                     -0.81 &                     -1.07 &                     -1.33 \\
FX         &              0.24 &                      0.13 &                      0.02 &                     -0.09 &                     -0.20 &                     -0.30 &                     -0.41 \\
Port.       &              \textbf{2.00} &                      \textbf{1.61} &                      \textbf{1.22} &                      \textbf{0.83} &                      0.44 &                      0.04 &                     -0.35 \\
\bottomrule
\end{tabular}

\end{table}
One of the weaknesses of DMNs is the performance net of transaction costs. This degradation is particularly acute in periods such as 2015--2020 where the model relies heavily on fast reversion.

In Exhibit \ref{tbl:cost-matrix-2015-2020} we detail the impact of transaction costs on the key architectures which we tested over 2015--2020. We increase the average cost $C$ from 0bps to 3bps, to give returns,
\begin{equation}
    \label{eqn:trans_cost}
    \bar{R}_{t+1}^{(i)} 
    = R_{t+1}^{(i)}
    - C\sigma_{\mathrm{tgt}}
    \left| \frac{ \mathbf{X}_t^{(i)}}{\sigma_t^{(i)}} - \frac{ \mathbf{X}_{t-1}^{(i)}}{\sigma_{t-1}^{(i)}} \right|.
\end{equation}
It is important to note that we still train the model on raw returns, however, for a proper treatment of transaction costs, we can directly account for this in the loss function with a turnover regulariser, as demonstrated in \cite{DeepMomentum}. To do this, we directly incorporate \eqref{eqn:trans_cost} into our loss function \eref{eqn:sharpe-loss} for optimum performance, given average transaction cost $C$.

We detail results by asset class, providing insight into the relative strengths and weaknesses of each architecture. Since there are more commodity futures in the portfolio, we look at the average Sharpe ratio for individual assets to ensure that we do not favour commodities which benefit more from diversification. When moving from raw returns to $C=3\text{bps}$, the LSTM experiences a total reduction of 228\%, whereas the Transformer is impacted the least and experiences a reduction of 106\%. This suggests that attention based architectures focus more on long term trends and therefore are less impacted by transaction costs. The Decoder-Only TFT also performs exceptionally well net of transaction costs, with a Sharpe ratio of 1.22 for $C=1\textrm{bps}$ when using a CPD module. The Transformer does outperform this architecture from $C=2\textrm{bps}$ onwards, which could be attributed to the fact that the TFT favours fast reversion more due to the LSTM component. Alternatively, this could also be because it has been optimised for $C=0\textrm{bps}$ and it was not focusing on transaction costs.

It can be noted that the LSTM performs well on equities and reasonably well on FX futures, which is likely due to its ability to learn localised patterns. The Transformer and Informer, on the other hand, perform well on the asset classes where it can identify longer trends, however, they perform poorly on FX where they need to be quicker. The Decoder-Only TFT gives the most rounded performance, performing very well across all asset classes, benefiting from both its LSTM and self-attention components. The addition of the CPD module has the biggest impact on commodities, where timing is particularly important. This translates to a superior portfolio performance, of which commodity futures is the biggest component. It should be noted that even at $C=3\textrm{bps}$ the model performs very well on commodities. The relatively weak performance on Fixed Income across all architectures could be attributed to the fact that the portfolio is light on Fixed Income futures and is not given much weighting during the training process. 

\section{Conclusions}
We have demonstrated that our attention-based model, the \textit{Momentum Transformer}, significantly outperforms the LSTM based \textit{Deep Momentum Network} (DMN), across all risk-adjusted performance metrics. We have illustrated the suitability of the \textit{Momentum Transformer} by back-testing over a sustained period of time from 1995--2020 and noting its comparatively strong performance in recent years. Our model is able to learn longer term patterns than the LSTM, benefiting from a longer input sequence length, specifically one year. Furthermore, all attention-based architectures, which we tested, are robust to significant events, such as during the SARS-CoV-2 market crash. Whilst an attention-LSTM hybrid Decoder-Only Temporal Fusion Transformer (TFT) model was the overall best performer, it can be noted that results from the non-hybrid architectures, such as the Transformer and Informer, are on an upward trend in recent years and actually outperformed the TFT model during the SARS-CoV-2 market crash. Nonetheless, we propose the TFT model because it is arguably more robust, performing well more broadly. However, due to the comparative strengths of each model depending on the asset class and regime, we suggest it could be worthwhile to use an ensembling approach, where multiple learning algorithms are utlised to obtain better predictive performance, if trading in practice.

Our detailed study of the results by asset class demonstrate that the \textit{Momentum Transformer} performs exceptionally well even net of costs. If we were to trade only with the 25 commodity futures in the period 2015--2020, we would still achieve a portfolio Sharpe ratio of 1.23 at average transaction cost $C=3\mathrm{bps}$. The reasoning for this could simply be because our portfolio contains more commodities than other asset classes, skewing its performance. It could be worthwhile to employ a transfer learning approach where we learn universal features \cite{UniveralFeatures}, which are not asset-specific, then update the model for each asset class.  

It is important to note that the \textit{Momentum Transformer} achieves impressive performance, only using price series information. An interesting avenue of future research would be expanding this work beyond futures to equities, incorporating factors such as Value and Quality. Here, we could take advantage of the larger universe of assets, fully utilising the deep-learning approach.


We deconstruct our deep-learning based momentum and mean-reversion strategy unlike any previous works. Our interpretable components help to shed light on how the model blends classical strategies based on the data. Looking at the interpretable attention patterns, we highlight the importance the model gives to significant events and how it segments the time-series into clearly defined regimes, learning regime-specific dynamics in the process. An interesting avenue for future work would be comparing this to Continual Learning, which is a paradigm whereby an agent sequentially learns new tasks. 

\section{Acknowledgements}
We would like to thank the Oxford-Man Institute of Quantitative Finance for financial and computing support. Furthermore, SR would like to thank the UK RAEng.


\newpage
\bibliographystyle{IEEEtran}
{\footnotesize
\bibliography{trading_with_the_momentum_transformer}
} 
\clearpage
\newpage
\appendix
\subsection{Dataset Details}
\label{apdx:data}

\begin{table}[!htb]
\centering
\caption{Portfolio assets \label{tbl:dataset}}
\begin{tabular}{lll}
\midrule
\textbf{Identifier} & \textbf{Description} & \textbf{\begin{tabular}[c]{@{}l@{}}Test     From\end{tabular}}       \\ \midrule
\multicolumn{2}{l}{{\underline{\textbf{Commodities (CM)}}}}\\
CC                  & COCOA                  & 1995    \\ 
DA                  & MILK III, composite      & 2000      \\
GI                  & GOLDMAN SAKS C. I.     & 1995    \\
JO                  & ORANGE JUICE       & 1995        \\
KC                  & COFFEE       & 1995              \\
KW                  & WHEAT, KC        & 1995          \\
LB                  & LUMBER       & 1995              \\
NR                  & ROUGH RICE      & 1995           \\
SB                  & SUGAR \#11    & 1995             \\
ZA                  & PALLADIUM, electronic  & 1995     \\
ZC                  & CORN, electronic   & 1995        \\
ZF                  & FEEDER CATTLE, electronic & 1995 \\
ZG                  & GOLD, electronic       & 1995    \\
ZH                  & HEATING OIL, electronic & 1995   \\
ZI                  & SILVER, electronic      & 1995   \\
ZK                  & COPPER, electronic      & 1995   \\
ZL                  & SOYBEAN OIL, electronic  & 1995  \\
ZN                  & NATURAL GAS, electronic  & 1995  \\
ZO                  & OATS, electronic         & 1995  \\
ZP                  & PLATINUM, electronic    & 1995   \\
ZR                  & ROUGH RICE, electronic   & 1995  \\
ZT                  & LIVE CATTLE, electronic  & 1995  \\
ZU                  & CRUDE OIL, electronic    & 1995  \\
ZW                  & WHEAT, electronic       & 1995   \\
ZZ                  & LEAN HOGS, electronic   & 1995  \\ \hline
\multicolumn{2}{l}{{\underline{\textbf{Equities (EQ)}}}}\\
CA                  & CAC40 INDEX       & 2000         \\
EN                  & NASDAQ, MINI        & 2005       \\
ER                  & RUSSELL 2000, MINI    & 2005     \\
ES                  & S\&P 500, MINI      & 2000     \\
LX                  & FTSE 100 INDEX        & 1995     \\
MD                  & S\&P 400 (Mini electronic) & 1995 \\
SC                  & S\&P 500, composite & 2000     \\
SP                  & S\&P 500, day session  & 1995  \\
XU                  & DOW JONES EUROSTOXX50   & 2005   \\
XX                  & DOW JONES STOXX 50   & 2005      \\
YM                  & Mini Dow Jones (\$5.00) & 2005   \\ \hline
\multicolumn{2}{l}{{\underline{\textbf{Fixed Income (FI)}}}}\\
DT                  & EURO BOND (BUND)  & 1995      \\
FB                  & T-NOTE, 5yr composite & 1995   \\
TY                  & T-NOTE, 10yr composite & 1995   \\
UB                  & EURO BOBL      & 2005          \\
US                  & T-BONDS, composite  & 1995     \\ \hline
\multicolumn{2}{l}{{\underline{\textbf{Foreign Exchange (FX)}}}}\\
AN & AUSTRALIAN \$\$, composite & 1995   \\
BN & BRITISH POUND, composite & 1995  \\
CN & CANADIAN \$\$, composite & 1995     \\
DX & US DOLLAR INDEX     & 1995       \\
FN & EURO, composite    & 1995        \\
JN & JAPANESE YEN, composite  & 1995  \\
MP & MEXICAN PESO     & 2000          \\
NK & NIKKEI INDEX  & 1995              \\
SN & SWISS FRANC, composite & 1995     \\ \bottomrule
\end{tabular}
\end{table}

We use the ratio adjusted continuous futures contracts from the Pinnacle Data Corp CLC Database \cite{PinnacleData}. Between each roll date, the contracts are multiplied by a fixed constant to eliminate jumps, which is computed moving backwards, starting from the current contract. This means that recalculation of the entire history is required at each roll date. All futures contracts in our portfolio are listed in Exhibit \ref{tbl:dataset} and have less than 10\% of data missing. We winsorise our data by limiting it to be within 5 times its exponentially weighted moving (EWM) standard deviations from its EWM average, using a 252-day half-life. This helps to limit the impact of outliers. We only use a contact if there is enough data available in the validation set for at least one input sequence. We list the start date of the fist out-of-sample training window in which we include a given asset in Exhibit \ref{tbl:dataset}. 

\subsection{Additional Details of Architectures}
\label{apdx:additional_details}

LSTMs \cite{lstm} were developed in response to the vanishing and exploding gradient problem \cite{vanishinggrad}, to help improve gradient flow. In addition to the output for each time-step $\mathbf{h}_t$, the hidden-state, The LSTM maintains $\mathbf{c}_t$, a cell state, which stores long-term information. The LSTM modulates information through a series of gates with $\mathbf{W}_{(\cdot)}\in\mathbb{R}^{m\times d_h}$ learnable weights, $\mathbf{U}_{(\cdot)}\in\mathbb{R}^{m\times d_h}$ learnable weights and $\mathbf{b}_{(\cdot)}\in\mathbb{R}^{d_h}$ learnable biases, for input $\mathbf{x}_t\in\mathbb{R}^{m}$ and hidden dimension $d_h$. The forget gate defines the information which can be ignored and the input gate determines the information which should enter the cell state. These two gates help the LSTM to handle non-stationarity via a dynamic autocovariance structure. The output gate helps to determine the information which flows to the next hidden state. We summarise each time-step of the LSTM as,
\begin{align}
\mathbf{f}_{t}&=\sigma (\mathbf{W}_{f}\mathbf{x}_{t}+\mathbf{U}_{f}\mathbf{h}_{t-1}+\mathbf{b}_{f}) &\text{(forget)}\\
\mathbf{i}_{t}&=\sigma (\mathbf{W}_{i}\mathbf{x}_{t}+\mathbf{U}_{i}\mathbf{h}_{t-1}+\mathbf{b}_{i})&\text{(input)}\\
\mathbf{o}_{t}&=\sigma (\mathbf{W}_{o}\mathbf{x}_{t}+\mathbf{U}_{o}\mathbf{h}_{t-1}+\mathbf{b}_{o})&\text{(output)}\\
{\tilde {\mathbf{c}}}_{t}&=\tanh(\mathbf{W}_{c}\mathbf{x}_{t}+\mathbf{U}_{c}\mathbf{h}_{t-1}+\mathbf{b}_{c})&\text{(cell)}\\
\mathbf{c}_{t}&=\mathbf{f}_{t}\odot \mathbf{c}_{t-1}+i_{t}\odot {\tilde{\mathbf{c}}}_{t}&\text{(cell)}\\
\mathbf{h}_{t}&=\mathbf{o}_{t}\odot \tanh(\mathbf{c}_{t})&\text{(hidden)}
\end{align}
with $\odot$ being the element-wise Hadamard product and $\sigma(\cdot)$ the sigmoid activation function. In this paper, we intialise with $\mathbf{h}_0=\mathbf{0}$ and $\mathbf{c}_0=\mathbf{0}$.

While RNN models capture the positional time-series pattern via their recurrent structure, the Transformer needs to preserve the positional context explicitly because the dot-product operation cannot capture this local context. We must inject some information about the relative or absolute position of the sequence items, which we add to the input embedding. Positional encoding can either be fixed or learnable, however, we find that there is little difference in the results in the context of momentum trading and it increases the number of model parameters unnecessarily. In this paper we use the standard positional encoding \cite{AttentionIsAllYouNeed} for day and we enrich the embedding with asset static information. In line with \cite{Informer}, we add a timestamp through learnable year and month embeddings. For the benchmark Transformer architectures, with the exclusion of the TFT, we construct our embedding as a sum of four separate parts 1) a scalar projection, of $m$ features to $d_q$, 2) the asset's entity embedding \cite{EntityEmbeddings}, 3) the local context, i.e. the timestamp, and 4) learnable year and month embeddings.  In general, it would also be possible to include an even finer timestamp granularity such as week of the year or day of the week. We summarise our $m$ model features, $\mathbf{u}^{(i)}_t\in\mathcal{U}$, for each time-step as,
    \begin{itemize}
        \item $\left\{r^{(i)}_{t-t',t}/\sigma_t^{(i)}\sqrt{t'} \,\vert\, t'\in\{1, 21,63,126,252\}\right\}$, as returns at different timescales,
        \item $\left\{\mathrm{M}_{t}^{(i)}(S,L) \vert (S,L) \in \mathcal{T}\right\}$, as MACD indicators with $\mathcal{T}=\left\{(8,24), (16,28), (32,96)\right\}$,
        \item $\left\{\nu^{(i)}_t(l), \gamma^{(i)}_t(l) \,\vert\, t'\in\{21, 126\}\right\}$, as changepoint severity and location at different timescales, if the CPD module is used.
    \end{itemize}
We convert $\mathbf{u}^{(i)}_t$ to an embedding vector $\mathbf{x}_t\in\mathbb{R}^{d_q}$ for each time-step, where we drop the $(i)$ superscript for brevity. We also use $t\in\{1,\ldots, T\}$ in this section, instead of $\{T-\tau+1,\ldots, T\}$, for simplicity.

\begin{figure}[tbph]
\centering
\includegraphics[width=0.60\linewidth]{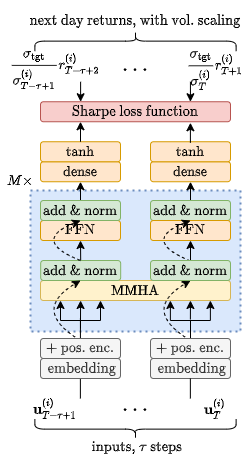}
\label{fig:dec-only-transformern}
\caption{Decoder-Only Transformer architecture. The components inside the dotted blue boxes can be stacked $M$ times.
}
\end{figure}

\begin{figure}[tbph]
\centering
\includegraphics[width=0.80\linewidth]{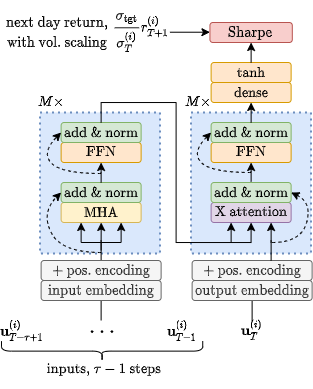}
\label{fig:enc-dec-transformern}
\caption{Encoder-Decoder Transformer architecture. It should be noted that, unlike the Decoder-Only architectures, only a single position is output rather than a series of positions corresponding to each input. Here, we can only compute the Sharpe loss function due to the other outputs in the mini-batch. The components inside the dotted blue boxes, for both the encoder and decoder, can be stacked $M$ times.
}
\end{figure}

The Feed-forward Network (FFN), which is used in all architectures, consists of a learnable linear transformation, or dense layer, followed by a Rectified Linear Unit (ReLU) activation function $\max(\cdot,\mathbf{0})$ \cite{DeepLearningBook}, to introduce non-linearity, then another learnable linear transformation.  We summarise the FFN as, 
\begin{equation}
    \mathrm{FFN}(\mathbf{x}_t) = \mathbf{W}_2\mathrm{ReLU}(\mathbf{W}_1 \mathbf{x}_t + \mathbf{b}_1) + \mathbf{b}_2,
\end{equation}
with parameters $\mathbf{W}_{(\cdot)} \in \mathbb{R}^{d_q \times d_q}$ and $\mathbf{b}_{(\cdot)} \in \mathbb{R}^{d_q}$, which we apply element-wise for each time-step $t$. 

In practice, we incorporate dropout \cite{dropout} $\delta(\cdot)$ into our Transformer, to help prevent overfitting. Furthermore, we employ residual connections around each component, helping the model to skip any unnecessary components, followed with layer normalisation \cite{AttentionIsAllYouNeed} $\phi(\cdot)$ to aid training, which normalises to zero mean and unit standard deviation. The full encoder, applied element-wise to $\mathbf{X}=(\mathbf{x}_t)^T_{t=1}$, is,
\begin{align}
    \mathbf{\mathbf{Y}} &= (\mathrm{Enc}_{M} \circ \mathrm{Enc}_{1})(\mathbf{X}), \\
    \mathrm{Enc}_i(\mathbf{X})  &= \phi(\mathbf{X}^\prime + \delta(\mathrm{FFN}(\mathbf{X}^\prime)), \\
    \mathbf{X}^\prime  &= \phi(\mathbf{X} + \delta(\mathrm{MHA}(\mathbf{X})),
\end{align}
where $\mathbf{Y}=(\mathbf{y}_t)_{t=1}^T$ is an abstract representation. The decoder, applied element-wise to output embedding, $\tilde{\mathbf{Z}}=(\tilde{\mathbf{z}}_t)_{t=1}^T$, is,
\begin{align}
    \mathbf{Z} &= (\mathrm{Dec}_{\mathbf{Y},M} \circ \mathrm{Dec}_{\mathbf{Y},1})(\tilde{\mathbf{Z}}) \\
    \mathrm{Dec}_{\mathbf{Y},i}(\tilde{\mathbf{Z}})  &= \phi(\tilde{\mathbf{Z}}^{\prime\prime} + \delta(\mathrm{FFN}(\tilde{\mathbf{Z}}^{\prime\prime})), \\
    \tilde{\mathbf{Z}}^{\prime\prime} &= \phi(\tilde{\mathbf{Z}}^{\prime} + \delta(\mathrm{XA}_\mathbf{Y}(\tilde{\mathbf{Z}}^{\prime})), \\
    \tilde{\mathbf{Z}}^{\prime} &= \phi(\tilde{\mathbf{Z}} + \delta(\mathrm{MMHA}(\tilde{\mathbf{Z}})).\label{eqn:mmha_dec}
\end{align}
In this paper we actually remove \eqref{eqn:mmha_dec} and set $\tilde{\mathbf{Z}}^{\prime} = \tilde{\mathbf{Z}}$, since we are only forecasting for a single time-step ahead, therefore no MMHA is required. 

Whilst \cite{EnhancingLocalityTS} suggested that the decoder side of the transformer architecture can be sufficient for time series forecasting, it is also argued by \cite{Informer} that Transformers are inherently designed as an encoder-decoder architecture and are superior to Decoder-Only Transformers. We test both the Transformer and Decoder-Only Transformer in this paper which we define as,
\begin{align}
    \mathbf{\mathbf{Z}} &= (\mathrm{DecO}_{M} \circ \mathrm{DecO}_{1})(\mathbf{X}), \\
    \mathrm{DecO}_i(\mathbf{X})  &= \phi(\mathbf{X}^\prime + \delta(\mathrm{FFN}(\mathbf{X}^\prime)), \\
    \mathbf{X}^\prime  &= \phi(\mathbf{X} + \delta(\mathrm{MMHA}(\mathbf{X})).\label{eqn:dec-only}
\end{align}
Unlike the full Transformer, we output a position for each time-step, $\mathbf{Z}=(\mathbf{z}_t)^T_{t=1}$, which avoids look-ahead bias with the MMHA step.

The TFT \cite{TFT} is constructed by piecing together a number of intelligent components, each with their own function, with the key components demonstrated in \eqref{eqn:tft_architecture}. In its original guise, it is an encoder-decoder architecture, however, for the \textit{Momentum Transformer} we propose a Decoder-Only variation. We borrow a number of the TFT components, including IMHA which we have previously detailed. The TFT is, an attention-LSTM hybrid model which uses recurrent LSTM layers for local processing and interpretable self-attention layers for long-term dependencies.  For the TFT, positional context is captured by the LSTM instead of explicitly adding positional encoding.

\begin{figure}[tbph]
\centering
\includegraphics[width=0.95\linewidth]{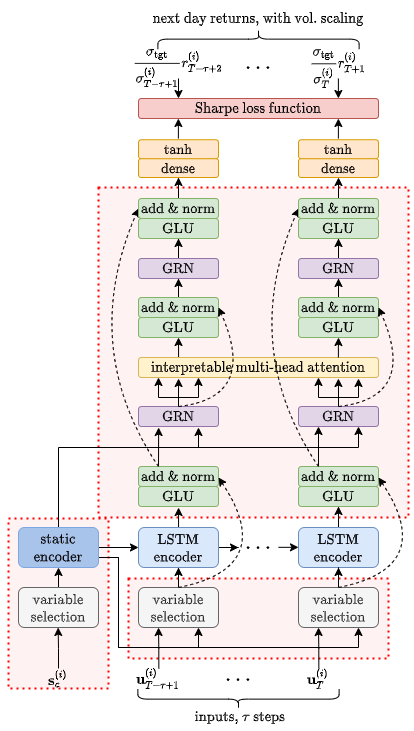}
\label{fig:momentum-transformern}
\caption{Decoder-Only TFT architecture which we refer to as the \textit{Momentum Transformer}. The dotted red boxes indicate the additional components we are adding to the LSTM-based DMN architecture.
}
\label{fig:architecture}
\end{figure}

The TFT supports static covariates, where Entity Embeddings \cite{EntityEmbeddings} are used as feature representations for categorical variables, and linear transformations for continuous variables.  In the context of the \textit{Momentum Transformer}, we include the asset type, 
\begin{equation}
    \mathbf{s}_c^{(i)} = (\textrm{asset}^{(i)}),
\end{equation}
which is a single categorical variable, encoding information for the asset-class. It is possible to include other static covariates here.

The Gated Linear Unit (GLU) \cite{GatedConvNetworks} is a component which we use to suppress any component which makes the architecture overly complex. For input $\mathbf{x}\in\mathbb{R}^{d_q}$, which dropout \cite{dropout} has been applied to,
\begin{equation}
    \mathrm{GLU}(\mathbf{x}) = (\mathbf{W}_1 \mathbf{x} + \mathbf{b}_1) \odot \sigma(\mathbf{W}_2 \mathbf{x} + \mathbf{b}_2).
\end{equation}
This is followed by an \textit{Add and Norm} component which adds the activation of the GLU with the output of a previous component followed with layer normalisation.

The \textit{Gated Residual Network} (GRN), proposed by \cite{TFT}, is a building block which applies non-linear processing, but only when required, to make our model more robust. In the case where we have a small or noisy dataset, this component defaults to a simpler linear model, with the key component being the \textit{Exponential Linear Unit} (ELU) \cite{ExponentialLinearUnits} which can either act as a linear or nonlinear layer. It is both preceded and followed by a dense layer. There is the ability to skip the building block via a GLU followed by an \textit{Add \& Norm}. Optionally, $(\mathbf{a},\mathbf{c}) \mapsto \mathrm{GRN}(\mathbf{a},\mathbf{c})$ can benefit from an additional static context input $s_c$ in addition to the primary information $\mathbf{a}$. 

This sample-dependent \textit{Variable Selection Network} \cite{TFT} (VSN) component is used to select the variables which are of most significance for the prediction problem, filtering out any inputs with a low signal rate. Weights are generated with a softmax after incorporating static information $\mathbf{s}_c\in \mathbb{R}^{d_q}$ and non-linear processing via a shared GRN,
\begin{equation}
    \eta(\tilde{\mathbf{x}}_{t,j}) = \frac{e^{\zeta_{t,j}}}{\sum^m_{i=1}e^{\zeta_{t,i}}}, \quad \zeta_{t,j} 
    = \mathrm{GRN}_{\zeta}(\tilde{\mathbf{x}}_{t,j}, \mathbf{s}_c).
\end{equation}
We apply the weights to a representation for each covariate, which involves an additional non-linear step via a GRN, $\mathrm{GRN}_{\psi_j}(\cdot)$, specific to each covariate,
\begin{equation}
    \mathbf{x}_t = \sum^{m}_{j=1}\eta(\tilde{\mathbf{x}}_{t,j})\mathrm{GRN}_{\psi_j}(\tilde{\mathbf{x}}_{t,j}).
\end{equation}
Explanation methods such as LIME \cite{LIME} and SHAP \cite{SHAP} can be applied post-hoc, providing insights into variable importance. However, unlike the VSN, these approaches fail to take into account time ordering. 

The Convolutional Transformer \cite{EnhancingLocalityTS}, is an extension to the Decoder-Only Transformer. It incorporates convolutional and log-sparse self-attention in order to increase the Decoder-Only Transformer’s awareness of locality as well as decreasing the quadratic memory cost of the attention-mechanism. It addresses the possibility that a single point in time might be very much dependent on the surrounding context and adds a causal convolution \cite{TimeSeriesDeepLearning} operation, of kernel size $k_c$ and stride 1, before the self-attention query-key.  An initial motivation was shopping patterns around holidays, however this concept fits in neatly with the concept of market regimes. Furthermore, the authors noted very little attention is typically given to most keys, and the attention mechanism typically focuses  on few important keys. The authors, therefore, proposed \textit{LogSparse} attention which only permits the model to attend to previous cells with an exponential step size, restricting the number of keys and parameters immensely.


The Informer, proposed by \cite{Informer}, is an extension to the Encoder-Decoder model, designed specifically for very long sequences. Again noting the sparsity of self-attention, the attention mechanism takes a probabilistic approach where the statistical distance, or Kullback–Leibler (KL) divergence \cite{kullback1951information}, 
\begin{equation}
  D_{\text{KL}}(Q\parallel P)=\sum _{\mathbf{x} \in S_\mathbf{X}} Q(\mathbf{x}) \log \left({\frac {Q(\mathbf{x})}{P(\mathbf{x})}}\right),
\end{equation}
between the the naive uniform distribution $Q$ and self-attention probability distribution $P$, as defined by \eqref{eqn:att_prob}, is calculated. The query is only set to active if there is substantial distance. This mechanism to distinguish essential queries is referred to as \textit{ProbSparse} self-attention. In this paper we set the top quartile of queries as active. Furthermore, the architecture decreases the dimension $d_q$ by half for each layer to further reduce the parameter space, which is a process termed as `distilling'.


\subsection{Experiment Settings}
\label{apdx:experiment_settings}
We calibrate our model using the training data by optimising on the Sharpe loss function via minibatch Stochastic Gradient Descent (SGD), using the \textit{Adam} optimiser \cite{ADAM}. We list the fixed model parameters for each architecture in Exhibit \ref{tab:fixed-params}. We keep the last 10\% of the training data, for each asset, as a validation set. We implement random grid search, as an outer optimisation loop, to select the best hyperparameters, based on the validation set. The parameter search grid for each architecture is listed in Exhibit \ref{tab:search-grid} and the number of search iterations is a fixed parameter. Rather than minimising validation loss, as in \cite{DeepMomentum, SlowMomFastRev}, we maximise the entire diversified strategy Sharpe ratio of the validation set. This helps to both stabilise the variability between repeated experiments and additionally leads to small improvements in risk-adjusted performance.  We implement early stopping, using the stopping patience listed for each architecture in Exhibit \ref{tab:fixed-params}.  Early stopping terminates training if there is no longer an increase in the Sharpe ratio of the validation set during this time period. Alternatively, we terminate training when the maximum number of epochs 

\begin{table*}[htbp]
\centering
\caption{Fixed Parameters}
\label{tab:fixed-params}
\begin{tabular}{lcccccc}
\toprule
\multicolumn{1}{c}{\multirow{2}{*}{\textbf{Parameters}}} & 
\multicolumn{1}{c}{\multirow{2}{*}{\textbf{LSTM}}} & 
\multicolumn{1}{c}{\multirow{2}{*}{\textbf{Transformer}}} &
\multicolumn{1}{c}{\textbf{Decoder-Only}} &
\multicolumn{1}{c}{\textbf{Convolutional}} &
\multicolumn{1}{c}{\multirow{2}{*}{\textbf{Informer}}} &
\multicolumn{1}{c}{\textbf{Decoder-Only}}
\\ 
& 
&
&
\multicolumn{1}{c}{\textbf{Transformer}} &
\multicolumn{1}{c}{\textbf{Transformer}} &
&
\multicolumn{1}{c}{\textbf{TFT}}
\\ \midrule
Sequence length, $\tau$ & 63 & 63 & 63 & 63 & 63 & 252 \\
$\Delta$ train start time & 63 & 1 & 63 & 63 & 1 & 252  \\
No. epochs & 300 & 50 & 300 & 300 & 50 & 300 \\
Stopping patience & 25 & 8 & 25 & 25 & 8 & 25\\
Search Iterations & 50  & 20 & 50  & 50  & 20  & 50\\
Train/valid ratio & 90\%/10\% & 90\%/10\%  & 90\%/10\%  & 90\%/10\%  & 90\%/10\%  & 90\%/10\%  \\
\bottomrule
\end{tabular}
\vspace{1ex}
\end{table*}

\begin{table*}[htbp]
\centering
\caption{Hyperparameter Search Grid}
\label{tab:search-grid}
\begin{tabular}{lcccccc}
\toprule
\multicolumn{1}{c}{\multirow{2}{*}{\textbf{Parameters}}} & 
\multicolumn{1}{c}{\multirow{2}{*}{\textbf{LSTM}}} & 
\multicolumn{1}{c}{\multirow{2}{*}{\textbf{Transformer}}} &
\multicolumn{1}{c}{\textbf{Decoder-Only}} &
\multicolumn{1}{c}{\textbf{Convolutional}} &
\multicolumn{1}{c}{\multirow{2}{*}{\textbf{Informer}}} &
\multicolumn{1}{c}{\textbf{Decoder-Only}}
\\ 
& 
&
&
\multicolumn{1}{c}{\textbf{Transformer}} &
\multicolumn{1}{c}{\textbf{Transformer}} &
&
\multicolumn{1}{c}{\textbf{TFT}}
\\ \midrule
Mini-batch size & 64, 128, 256 & 512, 1024 &  64, 128 & 64, 128 & 512, 1024 & 32, 64, 128\\\noalign{\vskip 1mm}  

Learning rate & 10\textsuperscript{-4}, 10\textsuperscript{-3}, 10\textsuperscript{-2}, 10\textsuperscript{-1}  &  10\textsuperscript{-4}, 10\textsuperscript{-3} & 10\textsuperscript{-4}, 10\textsuperscript{-3} & 10\textsuperscript{-4}, 10\textsuperscript{-3} & 10\textsuperscript{-4}, 10\textsuperscript{-3} & 10\textsuperscript{-4}, 10\textsuperscript{-3}, 10\textsuperscript{-2}, 10\textsuperscript{-1}\\\noalign{\vskip 1mm}
 
\multirow{2}{*}{Dropout} & 0.1, 0.2, 0.3, & 0.1, 0.2, 0.3, & 0.1, 0.2, 0.3, & 0.1, 0.2, 0.3, & 0.1, 0.2, 0.3, & 0.1, 0.2, 0.3, \\
 &  0.4, 0.5 & 0.4, 0.5 & 0.4, 0.5 & 0.4, 0.5 & 0.4, 0.5 & 0.4, 0.5 \\\noalign{\vskip 1mm}

Max grad. norm & 10\textsuperscript{-2}, 10\textsuperscript{0}, 10\textsuperscript{2}  &  10\textsuperscript{-3}, 10\textsuperscript{-2} , 10\textsuperscript{-1} & 10\textsuperscript{-3}, 10\textsuperscript{-2} , 10\textsuperscript{-1} & 10\textsuperscript{-3}, 10\textsuperscript{-2} , 10\textsuperscript{-1} & 10\textsuperscript{-3}, 10\textsuperscript{-2} , 10\textsuperscript{-1} & 10\textsuperscript{-2}, 10\textsuperscript{0}, 10\textsuperscript{2}\\\noalign{\vskip 1mm}

LSTM hidden & 5, 10, 20, & \multirow{2}{*}{-} & \multirow{2}{*}{-} & \multirow{2}{*}{-} & \multirow{2}{*}{-} & 5, 10, 20, \\
layer size, $d_h$ & 40, 80, 160 &  &  &  &  & 40, 80, 160 \\\noalign{\vskip 1mm}

No. heads, $H$ & - & 2,4 & 2,4 & 2,4,8 & 2,4,8 & 4 \\\noalign{\vskip 1mm}

No. layers, $M$ & - & 1, 2, 3 & 1, 2, 3 & 2, 3, 4 & 1, 2, 3 & - \\\noalign{\vskip 1mm}

Dimension, $d_q$ & - & 8, 16, 32, 64 & 8, 16, 32, 64 & 8, 16, 32, 64 & 8, 16, 32, 64 & $d_h$ \\\noalign{\vskip 1mm}

$d_q/d_\text{att}$ & - & 1, 2, 4, 8 & 1, 2, 4, 8 & 1, 2, 4, 8 & 1, 2, 4, 8 & - \\\noalign{\vskip 1mm}
Convolution, $k_c$ & - & - & - & 1, 3, 6, 9 & - & - \\\noalign{\vskip 1mm}

\bottomrule
\end{tabular}
\vspace{1ex}
\end{table*}

The LSTM and TFT models were implemented via the \texttt{Keras} API in \texttt{TensorFlow} \cite{tensorflow}. The other Transformer models were implemented in \texttt{PyTorch} \cite{pytorch} because the original implementations of the Informer and Convolutional Transformer were in this framework. We partition our training set into non-overlapping sequences for most architectures. The Transformer and Informer, however, benefited from training with overlapping sequences, where the subsequent sequence starts one time-step after the beginning of the previous sequence. We shuffled the order in which each sequence appears in an epoch. Details specific to each of the candidate architectures can be found in \cite{AttentionIsAllYouNeed, EnhancingLocalityTS, Informer, TFT} and further details of the implementation of DMNs can be found in \cite{DeepMomentum, SlowMomFastRev}.

\subsection{Additional Discussion} \label{apdx:additional_discussion}
In Exhibit \ref{fig:variable-importance} we can see how the variable importance for trading on Cocoa futures changes over time, where our model intelligently blends different strategies at different points in time, noting a change in strategy after significant events. In the lack of CPD, we observe the model take an approach where a number of strategies are blended, in this case initially with significant weighing on the MACD $\mathrm{M}_{t}^{(i)}(8,24)$ indicator and some weighting on monthly returns throughout the entire five year period.  Once the price drops in about mid 2016, significant importance is allocated to daily returns. The model continues to place high importance on daily returns as it is entering a mean-reverting regime, however once the price starts to rise again near the start of 2018, we shift back towards the original strategy. Once the price crashes for a second time, and we move back into a mean-reverting regime, daily returns become more important again. If the CPD features are added, the model adopts an entirely different strategy where it primarily uses knowledge from the 21 day CPD module to trade in conjunction with annual returns, shifting to returns with shorter timescales after any momentum turning points.

\end{document}